\documentclass[cameraready]{Interspeech}
\usepackage{booktabs}
\usepackage{tabularx}
\usepackage{multirow}
\usepackage{fvextra} % extends fancyvrb
\usepackage{url}
\usepackage{hyperref}
\usepackage[most]{tcolorbox}
\usepackage[table]{xcolor}
\usepackage{colortbl}     
\usepackage{graphicx}
\usepackage{caption} % for \captionof
\usepackage{tcolorbox}
\tcbuselibrary{breakable,skins}

\title{StanceBench: A Benchmark for Audio LLM-Based Interpersonal Stance Evaluation from Speech}

\author[affiliation={1}, orcid=0009-0008-4451-1319, correspondingauthor]{Yuzhe}{Wang}
\author[affiliation={1}, orcid=0000-0001-8953-7872]{Thomas}{Thebaud}
\author[affiliation={2}, orcid=0000-0003-4075-6876]{Jennifer}{Hu}
\author[affiliation={1}, orcid=0000-0001-9459-8426]{Jesús}{Villalba-Lopez}
\author[affiliation={3}, orcid=0009-0001-7214-2919]{Venkatesh}{Ravichandran}
\author[affiliation={4}, orcid=0000-0002-9910-6598]{Georgi}{Tinchev}
\author[affiliation={1}, orcid=0000-0002-4489-5753]{Najim}{Dehak}
\author[affiliation={1}, orcid=0000-0002-3033-7005]{Laureano}{Moro-Velázquez}
\address{
    $^1$ Electrical and Computer Engineering Department, Johns Hopkins University, Baltimore, MD, USA \\
    $^2$ Department of Cognitive Science, Johns Hopkins University, Baltimore, MD, USA\\
    $^3$ Amazon AGI, USA \\
    $^4$ Amazon Research, UK
}

\email{ywang792@jhu.edu}


\usepackage{comment}

\begin{document}

\maketitle

\begin{abstract}
% Speech-to-speech dialogue models increasingly rely on prosody and interaction nuance to convey social intent, yet evaluation still centers on content correctness. 
%We introduce a benchmark framework to quantify interpersonal stance in conversational speech and benchmark audio-aware LLM judges. 
% Speech-to-speech dialogue models increasingly rely on prosody and interaction nuance to convey social intent, but benchmark coverage of these signals remains limited. We introduce StanceBench, a benchmark framework that measures interpersonal stance in conversational speech and evaluates audio-capable LLMs as automated judges. Using the Seamless Interaction corpus, this framework (i) defines 10 stance dimensions from role-prompt poles, (ii) standardizes single-speaker and interaction-based evaluation pipelines, and (iii) reports LLM-as-a-judge robustness, biases, and stance inference abilities.  The evaluated commercial models outperform open-weight models. GPT yields the best overall separability in judging warmth, empathy, politeness, assertiveness, honesty, and attentiveness (AUROC up to 0.94). Gemini leads on empathy (0.96) and dominance/deference (0.86) judging. Among open-weight models, Qwen has less positional bias but has less separability; Kimi has strong separability but is the least robust in outputing structured output. 
Speech-to-speech dialogue models increasingly depend on prosody and interactional nuance to convey social intent, yet benchmarks for these cues remain limited. We introduce StanceBench, a benchmark for measuring interpersonal stance in conversational speech and evaluating audio-capable LLMs as automated judges. Using the Seamless Interaction corpus, StanceBench (1) specifies 9 stance dimensions via role-prompt poles, (2) standardizes single-speaker and interaction-based evaluations, and (3) reports LLM-as-a-judge robustness, bias, and stance inference. Across evaluated stances, empathy and politeness are the easiest. Warmth and assertiveness are moderately separable with positivity skew/asymmetry. Honesty is the hardest and shows high prompt order bias, consistent with needing cross-turn evidence. Attentiveness is separable but aligns weakly with humans. Interaction stances are more context-sensitive, with threshold gaps and high variance, especially conflict regulation.

% The analyzed commercial models outperform open-weight models across both settings overall. GPT has the strongest separability for warmth, empathy, politeness, assertiveness, honesty, and attentiveness (AUROC up to 0.94); Gemini leads on empathy (0.96) and dominance (0.86). Among open-weight models, Qwen shows lower positional bias but weaker separability. Kimi has strong separability but is least robust in producing structured outputs.
\end{abstract}

\section{Introduction}

Speech is a central medium for human interaction and communicates more than lexical content, conveying prosody, voice quality, and other paralinguistic cues that shape affect and social intent. Recent progress in speech-to-speech (S2S) modeling has shifted systems from cascaded pipelines toward end-to-end Speech Language Models (SpeechLMs) that process and generate speech, aiming to reduce information loss and latency while improving conversational naturalness \cite{cui2025recent}. Spoken dialogue modeling has likewise moved toward interaction settings where models operate on audio or audio-visual inputs and produce spoken responses without intermediate text as the primary representation \cite{park2024let}. However, evaluation has not kept pace: traditional ASR/TTS metrics prioritize transcription fidelity, intelligibility, and naturalness, but do not directly capture interaction-level behaviors that dominate user perception in dialogue, such as empathy, engagement, politeness, and dominance. Even in text-based dialogue, automatic metrics struggle to reflect these interactional qualities \cite{mehri2020usr}, and comprehensive evaluation of social and interactional meaning in speech remains limited \cite{ao2024sd}.

\begin{figure}[!t]
    \centering
    \includegraphics[width=1\linewidth]{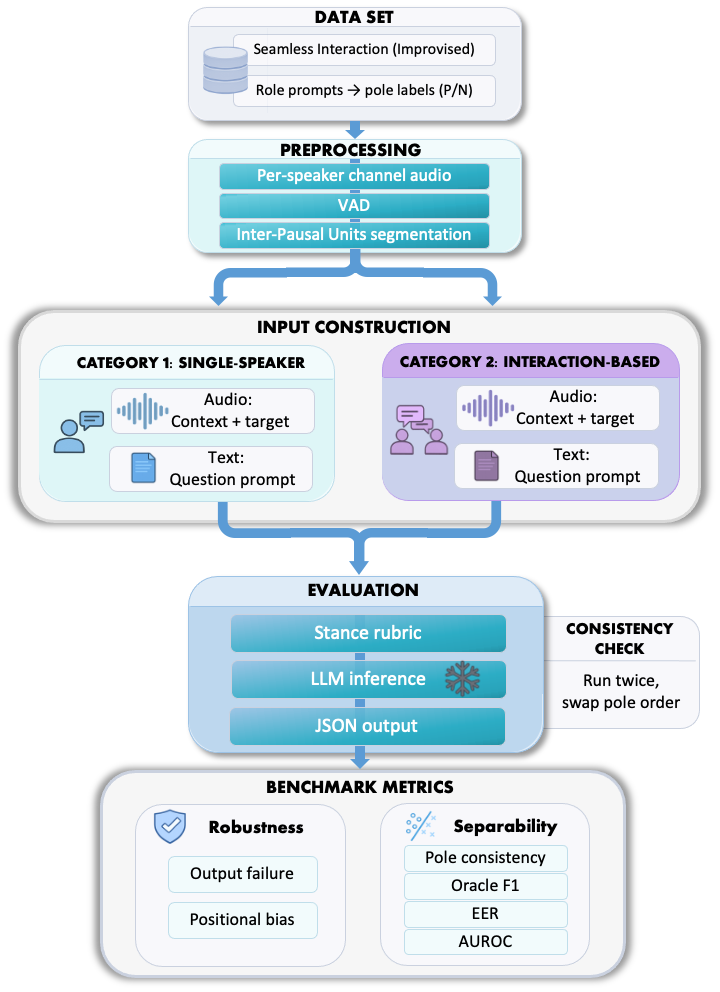}
    \caption{StanceBench evaluation pipeline. The pipeline begins with role-prompted conversations, followed by audio preprocessing and input construction for two evaluation settings. Audio segments are evaluated by an LLM judge under a stance rubric with a pole-order consistency check, and the resulting outputs are aggregated into robustness and separability metrics.}
    \label{fig:diagram}
    \vspace{-1.5em}
\end{figure}

\subsection{Interpersonal stance as an evaluation target}

In this study, we utilize interpersonal stance as a speaker’s socially meaningful positioning toward an interlocutor as expressed through acoustic delivery (e.g., prosody, tone) and interactional behavior (e.g., dominance, engagement), and we quantify it via rubric-based pole contrasts (e.g., warm/affiliative vs. cold/detached). Our goal is to standardize the evaluation of these stance signals in conversational speech, and to measure when audio-aware judges succeed or fail under a controlled pipeline.

Interpersonal stance is a particularly challenging target for automatic evaluation. Stance is expressed through an interaction between lexical choices, prosody, and conversational timing, and it is inherently context-dependent \cite{kuccuk2020stance}. Empirical evidence shows that stance-taking behaviors measurably affect turn-taking timing in dyadic conversation, reinforcing that stance is entangled with interaction dynamics rather than being a static label applied to isolated utterances \cite{ng2024investigating}. In S2S settings, this implies that stance evaluation must account for partner context and audio delivery, and it must tolerate ambiguity where multiple stances can be defensible. This also raises an immediate label-validity concern: because stance is perceptual and context-sensitive, “\textit{ground truth}” is difficult to define. Here we leverage the Improvised subset of the Meta Seamless Interaction corpus \cite{agrawal2025seamless}, where paired role prompts (mostly enacted by professional actors) explicitly elicit contrasting stance-like behaviors. We treat these role prompts as transparent intended stance labels (as weak supervision), and evaluate how well judges recover these intended contrasts.

\subsection{LLM-as-a-judge for speech evaluation}

A practical solution is the growing use of large language models as automated evaluators, often referred to as LLM-as-a-judge \cite{wang2025improving}. Given a rubric, a judge model can summarize evidence, assign scores, and generate a human-readable rationale. On certain tasks judge models can approximate human preference rankings when prompts are well specified \cite{bavaresco2025llms}. Judges also support structured, auditable outputs that are easy to aggregate and reuse across tasks, enabling rapid iteration against fine-grained rubrics \cite{li2025generation}. This approach is increasingly used in SpeechLMs evaluation: Pulikodan et al.\ introduced an LLM-based judging metric for ASR output quality in LLM-driven applications \cite{pulikodan2025approach}, and Chiang et al.\ explored audio-aware LLM judging for speaking style evaluation in generated speech \cite{chiang2025audio}. These studies support the feasibility of model-based evaluation for speech, but they primarily focus on content correctness or broad style control and leave open the need for a benchmark that operationalizes interpersonal stance as rubric-based pole contrasts in conversational speech and stress-tests audio-capable judges for robustness under controlled perturbations, a setting distinct from utterance-level emotion or empathy label prediction \cite{chen2024detecting}, fixed-label paralinguistic challenge benchmarks \cite{schuller2013interspeech}, and spoken dialogue evaluation suites such as SD-Eval \cite{ao2024sd}.

\subsection{StanceBench and contributions}

To address this gap, we introduce StanceBench, a benchmark-driven framework for interpersonal stance evaluation from speech that benchmarks candidate judge models under a unified, model-agnostic pipeline. StanceBench defines 9 role-prompt pole dimensions, standardizes single-speaker and interaction-based judging protocols, and reports robustness and stance separability with baseline results across judge models. This article makes three main contributions: (1) we propose a unified evaluation framework and benchmark suite for fine-grained speech assessment, emphasizing interpersonal stance recognition and stance-relevant interactional signals in speech; (2) we benchmark and analyze judge models themselves, quantifying reliability and limitations across stance-centric evaluation dimensions under a standardized evaluation pipeline; and (3) we identify a strong candidate judge under this role-derived labeling scheme and robustness tests, for future evaluation of S2S dialogue models. The complete code base and data will be released in a public repository.
\footnote{Code and benchmark resources are available at: https://github.com/YuzheWangjhu/StanceBench}

\section{Materials}

\subsection{Seamless Interaction Dataset}
This work uses the Seamless Interaction dataset~\cite{agrawal2025seamless}, a
large-scale corpus of in-person, dyadic interactions designed to
study social communication. The current release contains over
4,000 hours of two-person interactions from over 4,000 participants,
organized into short (2--10 min) conversations anchored by prompt pairs.
It spans both naturalistic conversations and prompted role-play
scenarios, providing broad coverage of conversational topics and
interpersonal stances.

We focus on the Improvised subset, in which participants (mostly actors) are assigned
explicit roles and instructed to speak under detailed prompts that
specify stance-like behaviors (e.g., affiliative vs.\ antagonistic;
deferential vs.\ dominant). Prompts were delivered separately to each
speaker to elicit targeted interpersonal dynamics that are rare or
difficult to capture reliably in unconstrained conversation.

Example role-prompt pair from the Improvised subset.:
\begin{itemize}
    \item Speaker A: (Devious) As the customer, approach the conversation with a sly and manipulative tone, trying to find loopholes or excuses to cancel your membership without penalty. Use persuasive language to make your case, while being prepared to bend the truth or make false promises to get what you want.
    \item Speaker B: (Approachable) As the gym owner, greet the customer warmly and listen to their concerns with empathy and understanding. Speak in a friendly and non-confrontational tone, asking questions to clarify their reasons for cancellation and offering alternative solutions or compromises to find a mutually beneficial outcome. Maintain a helpful and solution-focused demeanor throughout the conversation.
\end{itemize}

This example shows the role-prompt format used in the Improvised subset. This structure makes Seamless Interaction well-suited to our analysis for two reasons. First, role-level prompt pairs provide an explicit manipulation
of interpersonal stance, offering a principled basis for defining evaluation
dimensions and expected contrasts between speakers. Second, the dataset's design yields many stance-rich interactions across diverse contexts, supporting reliability analyses over a broad distribution rather than a small curated set. 

% We evaluate on 25\% of randomly sampled conversations from the Improvised subset
% due to runtime and computational resource limitations. A fixed seeding is used for all models to make sure they are evaluated on an identical subset. A total of 2431 conversations from 484 unique speakers are sampled. Per-dimension role-to-pole mappings are defined in Sec.~\ref{sec:dimensions}.
% We treat role prompts as intended stance labels (weak supervision) and report separability under this assumption.

\subsection{Judge Models}
We benchmark five judge models spanning open-weight and closed-source commercial
systems, covering end-to-end and cascaded pipelines as well as omni-modal and
audio-specialized architectures. All models act as evaluators under the same
stance rubric and a shared input construction pipeline; only the model-specific
inference adapter differs across runs.

\textbf{Qwen2.5-Omni-7B (Qwen)}
An end-to-end omni-modal model designed to perceive text, images, audio, and video,
while generating text and natural speech responses in a streaming manner. It adopts
a Thinker-Talker architecture that decouples text reasoning from speech generation,
serving as a compact open-weight baseline for audio-conditioned judging.
(\textbf{Checkpoint:} \texttt{Qwen/Qwen2.5-Omni-7B})~\cite{xu2025qwen25omnitechnicalreport}.

\textbf{Kimi-Audio-7B-Instruct (Kimi)}
An open-weight audio foundation model targeting unified audio understanding,
generation, and conversational interaction. It uses an audio tokenizer and an
audio LLM that maps audio features to discrete tokens, and is designed for a broad
range of audio-language tasks, making it an audio-specialized 7B judge in our benchmark.
(\textbf{Checkpoint:} \texttt{moonshotai/Kimi-Audio-7B-Instruct})~\cite{ding2025kimi}.

\textbf{ibm-granite/granite-speech-3.3-8b (Granite)}
We include a cascaded baseline that evaluates stance from \emph{transcript-based}
representations rather than end-to-end audio judging. Specifically, we use
Granite as the speech component and apply the same
rubric to its text representation.
(\textbf{Checkpoint:} \texttt{ibm-granite/granite-speech-3.3-8b})~\cite{saon2025granite}.

\textbf{gpt-audio (GPT)}
OpenAI's generally available audio model for the Chat Completions API, supporting both
audio inputs and audio outputs. We use \texttt{gpt-audio} as a closed-source, audio-native
judge to evaluate stance directly from speech signals.
(\textbf{Model ID/date:} \texttt{gpt-audio-2025-08-28})~\cite{openai_gpt_audio_model}.

\textbf{Gemini-2.5-Flash (Gemini)}
Google's multimodal model is positioned for strong price--performance tradeoffs,
described as a hybrid reasoning model with a controllable thinking budget. We include
Gemini-2.5-Flash as a widely deployed API judge that supports multimodal evaluation
settings and serves as a closed-source reference point.
(\textbf{Thinking budget:} \texttt{128}; \textbf{Model ID/date:} \texttt{gemini-2.5-flash/latest update June 2025})~\cite{google_gemini_25_flash_modelpage}.

% We evaluate ten interpersonal stance dimensions (S~0--S~8) derived from role prompts in the Seamless Interaction Improvised subset. For each dimension, we convert the dataset's role labels into a binary target by grouping semantically aligned roles into positive (P) and negative (N) poles. These groupings are grounded in established interpersonal and personality frameworks, including the Interpersonal Circumplex \cite{wiggins1979psychological, wiggins1995interpersonal}, the Big Five Aspect Scales structure \cite{deyoung2007between}, the HEXACO framework \cite{ashton2007empirical}, and social cognitive study \cite{fiske2007universal}. Role selections were made jointly by these literature-based criteria and by practical coverage considerations, to mitigate extreme pole imbalance. We evaluate on 25\% of randomly sampled conversations from the entire Improvised dateset due to runtime and computational resource limitations.

\section{Methods}

\subsection{Stance Dimensions}
% \subsubsection{Stance dimension construction}
\label{subsec:dimensions}
We evaluate 9 interpersonal stance dimensions (S~0--S~8) derived from role prompts in the Seamless Interaction Improvised subset.
% For each dimension, we convert the dataset’s role labels into a binary target by grouping semantically aligned roles into positive (P) and negative (\texttt{N}) poles. 
For each \textbf{stance dimension S}, we define two poles, \textbf{positive (\texttt{P})} and \textbf{negative (\texttt{N})}, and assign each role from the Seamless Interaction prompts to its intended pole. 
For example, for a certain conversation, if the prompt for one of the speakers had an \textit{Approachable} role, that speaker channel was mapped to \texttt{P} under S~0 dimension, \textit{Interpersonal Warmth}. If the role was \textit{Aloof}, it  was mapped to \texttt{N}.
These mappings are grounded in established interpersonal and personality frameworks, including the Interpersonal
Circumplex~\cite{wiggins1979psychological,wiggins1995interpersonal}, the Big Five Aspect Scales structure~\cite{deyoung2007between},
the HEXACO framework~\cite{ashton2007empirical}, and social cognitive study~\cite{fiske2007universal}.
Role selections were made jointly by these literature-based criteria and by practical coverage considerations. 

Because the Seamless Interaction role inventory contains many candidate roles for each defined stance dimension, we restrict each pole to a small representative subset (4 roles per pole, or 3 when fewer are available) to keep evaluation computationally tractable while preserving semantic coverage. Specifically, roles were chosen to semantically span the full range of their corresponding pole, represent its average characteristics, and mitigate extreme pole imbalances.

We evaluate all judge models on 25\% of randomly sampled conversations from the Improvised subset due to runtime
and computational constraints. We draw this subset once using a fixed random seed to ensure identical evaluation subset for every judge model. This yields 2431 conversations from 484 unique speakers.
We treat the role-prompt pole assignment as a weak label and report separability under this assumption.

% \subsubsection{Per-dimension role poles and judge questions}
% \label{subsubsec:Qdef}

Here we introduce the 9 dimensions assessed in this paper and the \texttt{P} and \texttt{N} roles associated with them. 
The following list details (in parentheses) the number of conversations per role and dimension, as well as the verbatim natural-language question asked to the LLM-judge relative to this dimension.
% All judge models share identical instances. 
% We report both threshold-free and thresholded separability metrics (Sec.~\ref{subsubsec:perfmancemetrics}) to account for imbalance.

\subsubsection{Category 1: Single-speaker evaluation (no interaction context)}
These dimensions are evaluated from a single speaker’s speech segment without the partner context.
% Category 1 segment (S~0--S~6)
\begin{itemize}

\item[\textbf{S~0}] \textbf{Interpersonal Warmth~\cite{wiggins1979psychological}}:\\
\textbf{\texttt{P} roles (127):} Welcoming; Approachable; Warm; Friendly\\
\textbf{\texttt{N} roles (101):} Indifferent; Impersonal; Distant; Aloof\\
\textbf{Question asked:} Based on the audio, does the TARGET speaker sound warm and affiliative, or cold and detached?

\item[\textbf{S~1}] \textbf{Compassion and Empathy~\cite{deyoung2007between}}:\\
\textbf{\texttt{P} roles (106):} Empathetic; Concerned; Understanding; Considerate\\
\textbf{\texttt{N} roles (192):} Insensitive; Callous; Inconsiderate; Unsympathetic\\
\textbf{Question asked:} Based on the audio, does the TARGET speaker sound compassionate and supportive, or callous and unsympathetic?

\item[\textbf{S~2}] \textbf{Politeness and Respect~\cite{deyoung2007between}}:\\
\textbf{\texttt{P} roles (120):} Polite; Respectful; Courteous\\
\textbf{\texttt{N} roles (88):} Disrespectful; Impolite; Uncivil\\
\textbf{Question asked:} Based on the audio, does the TARGET speaker sound polite and respectful, or rude and disrespectful?

\item[\textbf{S~3}] \textbf{Assertiveness~\cite{wiggins1979psychological}}:\\
\textbf{\texttt{P} roles (109):} Assertive; Decisive; Self-assured; Firm\\
\textbf{\texttt{N} roles (89):} Indecisive; Self-doubting; Unassertive; Timid\\
\textbf{Question asked:} Based on the audio, does the TARGET speaker sound assertive and self-confident, or hesitant and inhibited?

\item[\textbf{S~4}] \textbf{Sincerity and Honesty~\cite{ashton2007empirical}}:\\
\textbf{\texttt{P} roles (123):} Honest; Ingenuous; Uncalculating\\
\textbf{\texttt{N} roles (165):} Manipulative; Calculating; Devious\\
\textbf{Question asked:} Based on the audio, does the TARGET speaker sound straightforward and sincere, or sly and manipulative?

\item[\textbf{S~5}] \textbf{Cognitive Attentiveness~\cite{fiske2007universal}}:\\
\textbf{\texttt{P} roles (103):} Alert; Attentive; Concentrating; Engaged\\
\textbf{\texttt{N} roles (73):} Bewildered; Distracted; Drowsy; Unfocused\\
\textbf{Question asked:} Based on the audio, does the TARGET speaker sound attentive and focused, or confused and distracted?

% \item[\textbf{S~6}] \textbf{Organization and Motivation~\cite{deyoung2007between}}:\\
% \textbf{\texttt{P} roles (95):} Goal-oriented; Driven; Self-disciplined; Organized\\
% \textbf{\texttt{N} roles (194):} Unmotivated; Disorganized; Inconsistent; Unproductive\\
% \textbf{Question asked:} Based on the audio, does the TARGET speaker sound organized and goal-driven, or disorganized and unmotivated?

\end{itemize}

\vspace{0.5em}
\subsubsection{Category 2: Interaction-based evaluation (partner context available)}
These dimensions are evaluated in the conversational setting where interactional context is relevant.

% Category 2 segment (S~6--S~8)
\begin{itemize}

\item[\textbf{S~6}] \textbf{Social Engagement~\cite{deyoung2007between}}:\\
\textbf{\texttt{P} roles (83):} Engaging; Sociable; Gregarious; Outgoing\\
\textbf{\texttt{N} roles (159):} Withdrawn; Disengaged; Reticent; Taciturn\\
\textbf{Question asked:} Based on the audio, does the TARGET speaker sound socially engaged and expressive with the other speaker, or withdrawn and disengaged?

\item[\textbf{S~7}] \textbf{Power Orientation~\cite{wiggins1979psychological}}:\\
\textbf{\texttt{P} roles (127):} Submissive; Meek; Yielding; Undemanding\\
\textbf{\texttt{N} roles (143):} Overbearing; Dominant; Domineering; Forceful\\
\textbf{Question asked:} Based on the audio, does the TARGET speaker accommodate and yield to the other speaker’s preferences, or do they try to control and dominate?

\item[\textbf{S~8}] \textbf{Conflict Regulation~\cite{wiggins1995interpersonal}}:\\
\textbf{\texttt{P} roles (76):} Stable; Steady; Unaggressive; Unargumentative\\
\textbf{\texttt{N} roles (175):} Aggressive; Cruel; Ruthless; Vindictive\\
\textbf{Question asked:} Based on the audio, does the TARGET speaker stay calm and avoid confrontation, or are they hostile and aggressive?

\end{itemize}

\vspace{0.5em}

\subsection{Judging pipelines}
We formulate each stance dimension (S~0--S~8) as a binary-choice question with \texttt{P} and \texttt{N} poles. For each evaluation segment, the judge selects the pole that better matches the TARGET speaker given the provided audio. In Category~1, the judge receives only the TARGET speaker segment. In Category~2, the judge receives an interlocutor context segment in addition to the TARGET segment, and is instructed to evaluate the TARGET speaker while using the context to disambiguate interactional intent. (see subsection \ref{subsubsec:input} for segment definition)

\noindent\textbf{Example question specification:}

\vspace{0.25em}

\noindent\fbox{%
\begin{minipage}{0.97\columnwidth}
{\scriptsize\ttfamily\raggedright
question: "Based on the audio, does the TARGET speaker sound warm and affiliative, or cold and detached?"\\
positive\_definition: "Warm/affiliative: the TARGET speaker sounds friendly, welcoming, pleasant, and kind."\\
negative\_definition: "Cold/detached: the TARGET speaker sounds aloof, impersonal, indifferent, and socially distant."
\par}
\end{minipage}%
}

\subsubsection{Audio input construction}
\label{subsubsec:input}
Inter-Pausal Units (IPUs) are continuous stretches of speech produced by a single speaker and bounded by silent pauses exceeding a minimum duration. They are used as fundamental, objective units for segmenting and analyzing spoken language \cite{arora2025talking}. We implement per-speaker IPU extraction with an energy-based hysteresis voice activity detection (25\,ms window, 10\,ms hop) on each channel independently using the following rules:
\begin{itemize}
  \item A silence gap of 0.3s splits adjacent speech into separate IPUs.
  \item IPUs with active speech $< 0.2s$ are marked as backchannels.
  \item IPU length is defined by active speech duration (not wall-clock duration).
\end{itemize}

\noindent\textbf{Category 1: single-speaker segments.}\\
% Each evaluation example is an audio segment from a single speaker. We exclude any single IPU with active speech $> 45s$ to avoid unusually long, low-information turns dominated by neutral prolonged speech. The remaining IPUs are then concatenated to form multiple segments per speaker:
% \begin{itemize}
%   \item Use a fixed inter-IPU boundary of 0.25s when concatenating IPUs.
%   \item Stop adding the next IPU if (i) active speech would exceed 45s, or (ii) the boundary fraction would exceed 25\%.
%   \item Enforce that all segments except the final tail segment have active speech in $[30, 45]$ seconds.
% \end{itemize}
Each evaluation example is a single-speaker audio segment constructed by merging IPUs from the same speaker. We use VAD-based \emph{active speech duration} for all speech-length thresholds.
\begin{itemize}
  \item Exclude backchannels and any single IPU whose active speech exceeds 45s, to avoid unusually long, low-information turns dominated by neutral prolonged speech.
  \item Concatenating the remaining IPUs per speaker while inserting a fixed 0.25s silence between adjacent IPUs.
  \item Stop adding the next IPU if either:
    \begin{itemize}
      \item active speech would exceed 45s, or
      \item the \emph{boundary fraction} would exceed 25\%, where boundary \\
      \vspace{0.05px}\\
      fraction is defined as {\footnotesize $\displaystyle \frac{\text{total inserted inter-IPU silences length}}{\text{current evaluation segment length}}$\%}.
    \end{itemize}
  \item To standardize evidence length across examples, all segments except the final tail segment must have a minimum length of 30s and maximum of 45s.
\end{itemize}

\noindent\textbf{Category 2: Interaction-based.}\\
Each dyad has two speakers ($A$ and $B$). A detected turn-switch $A$$\rightarrow$$B$ means $A$ is the active speaker immediately before the turn switch and $B$ is the active speaker immediately after. For each turn-switch, we create two directional evaluation examples: \textbf{post\_target} evaluates $B$ as TARGET using $A$ as CONTEXT, and \textbf{pre\_target} evaluates $A$ as TARGET using $B$ as CONTEXT. 
In both cases, the judge is instructed to evaluate only the TARGET speaker while using CONTEXT to disambiguate interactional intent. The construction pipeline is:
\begin{enumerate}
  \item Detect turn switches and identify the preceding speaker and following speaker ($A$$\rightarrow$$B$ or $B$$\rightarrow$$A$).
  \item For each boundary, initialize a pre-turn-switch side window (closest IPU before the switch) and a post-turn-switch side window (closest IPU after the switch).
  \item Expand the pre- and post-turn-switch windows by adding adjacent IPUs from the same speaker until the CONTEXT has $\ge 1$s active speech and TARGET has $\ge 2$s active speech, or until reaching the limits (at most 6 added IPUs and within 30s wall-clock of the turn-switch on that side). To ensure the extracted audio stays local to the turn-switch, we only add IPUs that are contiguous with the switch. Specifically, if the next candidate IPU is separated from the turn switch by an IPU from the other speaker containing $\ge 0.8$s of active speech (i.e., the interlocutor produced a non-trivial intervening turn), we stop expanding on that side rather than ``jumping over'' the partner turn to include more temporally distant speech.
  \item Construct CONTEXT by concatenating the context-side IPUs with 0.25s gaps, then trim by \emph{active speech} to $[1,10]$ seconds: if CONTEXT is pre-turn-switch, keep the \emph{last} 10s of active speech; if CONTEXT is post-turn-switch, keep the \emph{first} 10s.
  \item Construct TARGET similarly, but trim by \emph{active speech} to $[2,15]$ seconds: if TARGET is pre-turn-switch, keep the \emph{last} 15s of active speech; if TARGET is post-turn-switch, keep the \emph{first} 15s.
\end{enumerate}
% We construct many context/target evaluation pairs centered on detected speaker turn switches. Each example is a pair of audio segments derived from a single turn switch boundary:
% \begin{itemize}
%   \item \textbf{post\_target}: the first segment from the two turns is context and the second, is the target to  be evaluated by the judge model.
%   \item \textbf{pre\_target}: the second turn is context and the first one is the target.
% \end{itemize}
% The pipeline is:
% \begin{enumerate}
%   \item Detect switch events (A$\rightarrow$B or B$\rightarrow$A).
%   \item For each switch, define a pre-side window (preceding IPUs) and a post-side window (following IPUs) around the boundary.
%   \item Expand both windows (up to 6 IPUs per side, with a maximum 30s lookback) until each side contains sufficient usable content.
%   \item Construct context audio: merge the context-side IPUs with 0.25s gaps, then trim by active speech to $[1, 10]$ seconds.
%   \item Construct target audio: merge the target-side IPUs with 0.25s gaps, then trim by active speech to $[2, 15]$ seconds.
% \end{enumerate}

\subsubsection{Judge models prompting}
Below we show the verbatim prompts used for both categories across all judge runs.

\label{fig:cat1prompt}
\begin{tcolorbox}[
  breakable,
  enhanced,
  sharp corners,
  colback=white,
  boxrule=\fboxrule,
  boxsep=\fboxsep,
  left=0pt,right=0pt,top=0pt,bottom=0pt,
  width=0.97\columnwidth,
  overlay first={\draw[black!25,line width=0.3pt] (frame.south west) -- (frame.south east);},
  before upper={\scriptsize\ttfamily\textbf{Category 1 system prompt (one audio: TARGET)}\par\medskip}
]
{\scriptsize\ttfamily\raggedright
You are a careful dialogue analyst. You will hear ONE audio segment spoken by the TARGET speaker. Your task is to judge the TARGET speaker's stance/tone/style relative to the question.\par\medskip
IMPORTANT OUTPUT FORMAT:\par
- Output ONLY one JSON object with keys 'choice', 'probability', and 'evidence'.\par
- 'choice' must be either "P" or "N". Always choose one.\par
- 'probability' must be a number from 0.0 to 1.0 representing probability that your choice is correct (0.0 to 1.0).\par
- If unsure, choose the closer option but set probability low.\par
- 'evidence' must be a JSON array of 2 to 4 short strings describing observable cues (tone, prosody, wording, engagement).\par
- No extra text.\par
Example JSONs (format only):\par
Example 1: \{ "choice":"P","probability":0.60,"evidence":[ "Cue about tone/prosody","Cue about wording" ] \}\par
Example 2: \{ "choice":"N","probability":0.60,"evidence":[ "Cue about tone/prosody","Cue about wording" ] \}\par
}
\end{tcolorbox}
\vspace{-0.5em}

\label{fig:cat2prompt}
\begin{tcolorbox}[
  breakable,
  enhanced,
  sharp corners,
  colback=white,
  colframe=black,
  boxrule=\fboxrule,
  boxsep=\fboxsep,
  left=0pt,right=0pt,top=0pt,bottom=0pt,
  width=0.97\columnwidth,
  before upper={\scriptsize\ttfamily\textbf{Category 2 system prompt (two audios: CONTEXT $\rightarrow$ TARGET; or swapped order TARGET $\rightarrow$ CONTEXT shown in parentheses)}\par\medskip}
]
{\scriptsize\ttfamily\raggedright
You are a careful dialogue analyst. You will hear TWO audio segments. The first audio is CONTEXT from the other speaker. The second audio is the TARGET speaker's response and is the only segment you should evaluate. (The first audio is the TARGET segment and is the only segment you should evaluate. The second audio is CONTEXT from the other speaker.) \par\medskip
IMPORTANT OUTPUT FORMAT:\par
- Output ONLY one JSON object with keys 'choice', 'probability', and 'evidence'.\par
- 'choice' must be either "P" or "N". Always choose one.\par
- 'probability' must be a number from 0.0 to 1.0 representing probability that your choice is correct (0.0 to 1.0).\par
- If unsure, choose the closer option but set probability low.\par
- 'evidence' must be a JSON array of 2 to 4 short strings describing observable cues in the TARGET speaker's response (tone, prosody, wording, engagement).\par
- Use the context only to interpret the TARGET response; do not judge the context speaker.\par
- No extra text.\par
Example JSONs (format only):\par
Example 1: \{ "choice":"P","probability":0.60,"evidence":[ "Cue about tone/prosody","Cue about wording" ] \}\par
Example 2: \{ "choice":"N","probability":0.60,"evidence":[ "Cue about tone/prosody","Cue about wording" ] \}\par
}
\end{tcolorbox}
\vspace{-0.5em}

\subsubsection{Judge output and consistency check}
After receiving the prompt and audio input, the judge is required to return a strict JSON object:
\\
\texttt{\{"choice":"P"|"N", "probability":0..1, "evidence":[2-4 strings]\}}\\
where \texttt{choice} indicates the selected pole, \texttt{probability} is the judge's confidence in the selected pole, and \texttt{evidence} contains 2--4 short textual cues supporting the decision. 

To probe positional sensitivity and improve robustness, we run the judge \textbf{twice} for every segment with swapped pole order:
\begin{itemize}
  \item \textbf{Variant 0}: \texttt{P} definition first, then \texttt{N}.
  \item \textbf{Variant 1}: \texttt{N} definition first, then \texttt{P}.
\end{itemize}
We run both orderings to probe positional sensitivity and mitigate position bias in LLM-based judging, which has been shown to affect evaluator fairness and stability \cite{wang2024large}.
We record both outputs and use their agreement/disagreement as a reliability signal in the downstream analysis.

All judge runs are deterministic with sampling disabled.
Each segment is attempted once per pole-order variant. If the output is malformed JSON or does not satisfy the required schema, the instance is discarded and counted toward the model’s failure rate. 

We explicitly request a verbal probability to elicit confidence estimates from the judge, following prior work on prompting strategies for confidence reporting~\cite{tian2023just}.
We do \emph{not} treat these probabilities as calibrated; they are used only as a continuous score to quantify the strength of the \texttt{P}/\texttt{N} selection.

\subsubsection{Text-only Ablation}
To isolate the contribution of acoustic information in stance judging, we compare two Qwen-based evaluation pipelines that differ only in whether the judge has access to raw audio. In the baseline setting, the Qwen judge model (default Qwen2.5-Omni-7B) receives prompts that include audio objects for the context and target segments, and the instructions allow evidence from both wording and paralinguistic cues such as tone and prosody. In the ablation setting, we add an ASR preprocessing stage using \texttt{gpt-4o-transcribe} to transcribe each context and target segment once, and we provide the Qwen judge with transcript-only text. Judgment is restricted to cues observable in the transcript.
% and instructs the model not to infer prosody beyond the words. 
All evaluation instances, Balanced-Position variants, scoring, and filtering procedures are otherwise held fixed across the two settings.

\subsection{Metrics}
\label{subsec:metrics}
For each stance dimension (S~0--S~8), we report (i) run-level robustness metrics that characterize judge stability, (ii) binary classification metrics that quantify stance separability under the role-group prompt-intended labels, and (iii) distributional diagnostics via \texttt{P} vs. \texttt{N} role-group violin plots.

\subsubsection{Metadata and run robustness}
To contextualize judge performance and support fair cross-model comparisons, we report \textbf{run-level robustness statistics} capturing dataset scale, evaluation coverage, and judge stability. For each question, we restrict analysis to the subset of conversation instances shared across all compared judge models. Concretely, we retain only those conversation instances for which \emph{every} model returns a valid structured judgment under the required output schema. This intersection avoids confounds from model-specific or stochastic failures (e.g., malformed outputs, tool errors, refusals) that may occur on arbitrary conversation instances. Additionally, a conversation instance is considered \emph{successful} for a model only if both pole-order variants return valid structured JSON under the required schema. Specifically, we report:
\begin{itemize}
  \item \textbf{Evaluated instances:} the number of conversation instances included after enforcing a common successful subset across all judge models.
  \item \textbf{Failure rate:} the fraction of attempted conversation evaluations that do not yield a valid judgment output under the required schema.
  \item \textbf{Mean segment disagreement rate:} the average rate at which the judge changes its decision between the two pole-ordering variants (Variant~0 vs.\ Variant~1), computed at the segment level and then aggregated over evaluated speakers.
\end{itemize}

We further exclude conversation instances where the evaluated model exhibits high positional bias. Specifically, if more than 20\% of segments within a conversation disagree between Variant~0 and Variant~1, this conversation is removed. All downstream metrics are computed on the resulting post-filter set.

\subsubsection{Score aggregation}
% Each segment is evaluated twice in Variant~0 and Variant~1. In each run, the judge returns a discrete pole decision (\texttt{P} or \texttt{N}) and an associated probability in $[0,1]$. We map each output to a \textbf{signed probability} by assigning a positive sign to \texttt{P} and a negative sign to \texttt{N}, producing an segment-level continuous signal in $[-1,1]$ that captures both \emph{direction} (the preferred pole) and \emph{probability} (the strength of support).

% For each segment, we compute the final score as the mean of the signed probabilities from Variant~0 and Variant~1, which reduces sensitivity to pole ordering. We then compute a conversation-level score by averaging segment scores across all available segments. The resulting continuous signal lies in $[-1,1]$, where larger values indicate stronger evidence for the positive pole and more negative values indicate stronger evidence for the negative pole.

Each segment is evaluated twice (Variant~0 and Variant~1). For each run, we convert the judge output to a signed probability in $[-1,1]$ by assigning $+p$ to \texttt{P} pole and $-p$ to \texttt{N} pole, where $p\in[0,1]$ is the reported probability. We define the segment score as the mean of the two signed probabilities, reducing sensitivity to pole ordering. We then compute a conversation-level score by averaging segment scores across all available segments. The resulting score lies in $[-1,1]$, with larger values indicating stronger evidence for the \texttt{P} pole and more negative values indicating stronger evidence for the \texttt{N} pole.

% \noindent\textbf{Decision rule (binary prediction).}
% We convert the continuous signed probability into a binary prediction by thresholding at zero: non-negative values are predicted as \texttt{P}, and negative values are predicted as \texttt{N}.

\subsubsection{Binary separability metrics}
\label{subsubsec:perfmancemetrics}
 Using the aggregated signed-probability signal $s \in[-1,1]$, we report four complementary metrics:
\begin{itemize}
  \item \textbf{Categorical pole-consistency:} we predict \texttt{P} if $s \ge 0$ and \texttt{N} otherwise. This metric computes the fraction of evaluated conversation instances for which the predicted pole matches the expected pole implied by the role grouping. This reflects performance under the benchmark's fixed decision rule (a single operating point) and enables direct cross-model comparison.

    \item \textbf{Oracle upper bound F1:} the maximum binary F1 obtained by sweeping a threshold $t$ on the conversation score $s$ (predict \texttt{P} if $s \ge t$, otherwise \texttt{N}) on the evaluated subset. We report this only to quantify the best achievable F1 if the threshold were perfectly tuned for that dimension and split, providing an upper bound on performance under optimal thresholding.

  \item \textbf{Equal Error Rate (EER):} the error rate at the threshold where false positive and false negative rates are equal. EER summarizes performance at a balanced operating point under symmetric error trade-offs.  
  
  \item \textbf{Area Under the Receiver Operating Characteristic Curve (AUROC):} computed from the same conversation scores $s$, it measures how well the judge signal separates \texttt{P}/\texttt{N}-role conversation instances across all possible thresholds, independent of any single operating point.

\end{itemize}

% Together, these metrics provide a more complete view than any single score. Categorical pole-consistency reflects performance under the \emph{fixed} decision rule used throughout the benchmark, directly measuring how often each judge makes the correct binary call in the standardized setting. AUROC instead evaluates the signed-probability signal as a \emph{ranking} function, helping distinguish poor separability from an ill-chosen threshold. EER adds an operating-point summary emphasizing \emph{balanced} trade-offs between false positives and false negatives. Finally, the oracle upper bound F1 serves as a separability ceiling, indicating whether a judge can produce a discriminative signal even if its default calibration or thresholding is imperfect. 
Together, the metrics disentangle decision accuracy, ranking quality, balanced error behavior, and threshold sensitivity.

  % We additionally provide a \textbf{human correlation} score that measures agreement with human judgments: for each dimension, we align model outputs with human consensus and compute Spearman’s correlation between the model score and the mean human score across participants, where 6 evaluators form two identical-within-group subsets (three per subset) whose union covers 122 unique conversation instances.
  %   Given the subset’s limited size and coverage, we treat human correlation as a preliminary validation test and will expand annotation scale and stratified coverage in future work.
\subsubsection{Human correlation}

We additionally report a \textbf{human correlation} score as a 
measurement of alignment between model judgments and human perception. For each stance dimension, we keep only instances where all five judge models produce valid outputs, and sample a small role-balanced subset with a fixed random seed for human evaluation (122 unique instances in total). 
Six human evaluators participated as two groups of three; within each group, all three evaluators annotated the same subset, yielding three ratings per instance. 
Annotators were shown prompts closely matching those given to the LLM judges, selected either the \texttt{P} or \texttt{N} pole, and provided a confidence score in $[0,1]$ in increments of $0.05$. 
Segment scores are first averaged within each annotator to obtain an instance-level score, and these instance-level scores are then averaged across annotators to obtain a human consensus score.
For each dimension, we compute Spearman’s correlation between the model’s instance-level score and the corresponding human consensus score. This measure
is used to assess model-human alignment on the sampled subset,
not to validate the role-prompt labels as perceived ground truth.

% \subsubsection{Distributional diagnostics: binary violin plots}
% To complement scalar performance metrics, we visualize pole separability using \textbf{two-group violin plots} for each model and stance dimension. For each model, we plot the distribution of aggregated signed probabilities in $[-1,1]$ for conversations in the positive-role set and the negative-role set. Each panel shows KDE-based density shapes for\texttt{P} and \texttt{N}, summary overlays (min/max and median), and jittered points for individual conversation instances. These plots provide an interpretable view of separability, probability distribution, and pole-specific asymmetries that may not be evident from a single aggregate score.

% To support visually meaningful cross-model comparisons, violin widths are normalized using a global density scale computed across all models for the same dimension. This ensures that violin thickness reflects absolute concentration, preventing similar-looking panels from obscuring substantial differences in distribution sharpness or sample concentration across models.

\subsubsection{Distributional diagnostics: violin plots}
To complement scalar performance metrics, we visualize pole separability using violin plots arranged as a model-by-pole grid. Rows correspond to judge models, and the two columns show conversations from the \texttt{P} and \texttt{N} poles, respectively. For each model and stance dimension, we plot the distribution of conversation-level signed-probability scores in $[-1,1]$. Each half-violin shows a density shape with median bar. This layout provides an interpretable view of separability, score concentration, and pole-specific asymmetries that may not be evident from a single aggregate score.

To support visually meaningful cross-model comparisons, for each stance dimension (S~0, S~1, etc.), we use one common width scale shared across all models and both poles within each dimension. This makes violin thickness directly comparable across models within that dimension.

\section{Results and Discussion}
Table~\ref{tab:judge_metrics_by_Q} summarizes StanceBench results for five candidate judges across 9 stance dimensions (S~0--8), reporting both
(i) \emph{judge robustness} under a strict structured-output contract (failure rate and pole-order sensitivity) and
(ii) \emph{stance separability} under role-derived poles (categorical pole-consistency, oracle F1, EER, AUROC).
Figure~\ref{fig:violin_plot} complements the scalar metrics with violin plots of the signed-probability scores for \texttt{P} vs.\ \texttt{N} role groups, showing distributional skew, overlap, and saturation effects that are easy to miss from a single number.
We separate reliability analysis from task separability analysis, since LLM-as-a-judge performance depends jointly on (a) output robustness and bias under prompt perturbations and (b) discriminative performance \cite{bavaresco2025llms,li2025generation,wang2024large}.

\subsection{Judge robustness}
Structured-output reliability differs by more than an order of magnitude across judge models.
The commercial API judges are consistently robust (GPT: 0.00--0.005 failure; Gemini: 0.00--0.02), and the cascaded Granite baseline is similarly stable (0.00--0.05).
In contrast, open end-to-end audio judges fail substantially more often (Qwen: 0.03--0.30; Kimi: 0.01--0.44, peaking at 0.44 on S~3).
Given our protocol (single attempt per pole-order variant; discard malformed JSON), these failures directly reduce usable coverage and are a first-order practical constraint.

Pole-order sensitivity is a distinct reliability axis related to position effects in LLM judging \cite{wang2024large}.
Qwen is highly order-stable on several single-speaker dimensions (0.01 on S~3 and S~5) but becomes unstable on S~8 (0.22).
Granite exhibits its largest order sensitivity on S~4 (0.29), while GPT’s most pronounced sensitivity occurs on power orientation (S~7: 0.18).
Overall, these results reinforce a key benchmark design point: high stability under pole-order perturbations does not guarantee strong separability, and vice versa. We therefore report both axes explicitly, rather than assuming evaluator competence.

\subsection{Stance separability}
% Stance separability measures how well role-derived poles induce a consistent separation between \texttt{P}/\texttt{N} groups (Figure~\ref{fig:violin_plot}).
% AUROC/EER evaluate score ranking, while categorical pole-consistency reflects a fixed decision rule; discrepancies between them indicate calibration or direction ambiguity.

\begin{figure}[!t]
    \centering
    \includegraphics[width=1\linewidth]{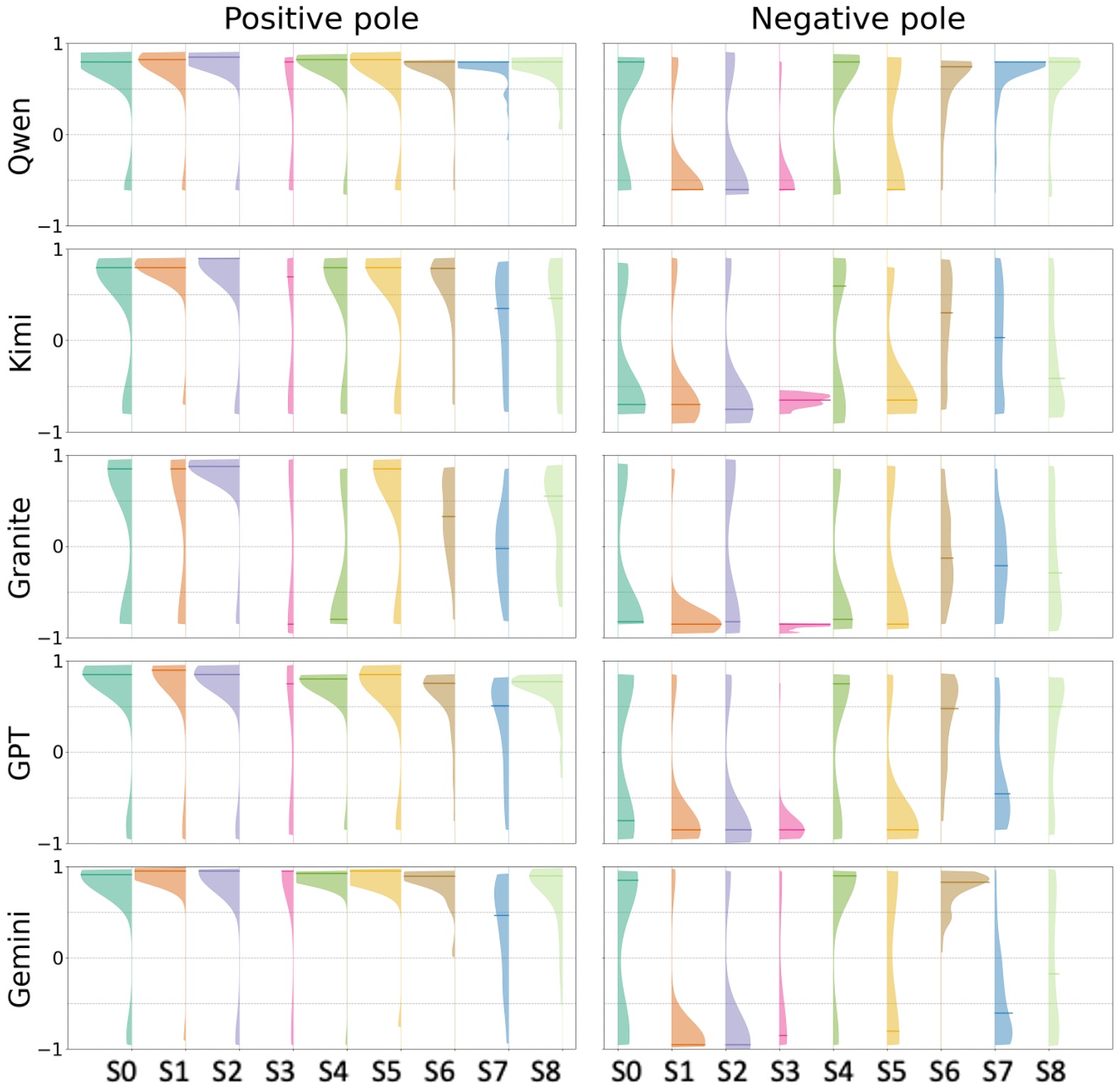}    
    \caption{Half-violin distributions of conversation-level signed-probability scores for judge models (rows) and stance dimension (x-axis). Median bars summarize central tendency. Widths are capped at the 95th percentile for visual interpretability.}
    % For each stance dimension, a common width scale is shared across all models and both pole columns to enable cross-model comparison.
    \label{fig:violin_plot}
    \vspace{-2em}
\end{figure}

\textbf{Warmth} (S~0) is only moderately separable (AUROC 0.66--0.81). Warmth is socially normative, so positive-pole scores often cluster near mild-positive, while \texttt{N} instances drift upward toward 0 in Figure~\ref{fig:violin_plot}, consistent with the negative pole often being expressed indirectly (flat prosody, short responses) rather than via a single strong cue. Human correlation is moderate (0.53--0.80), suggesting judges track human warmth rankings but some still default toward mild-positive when evidence is subtle.

\textbf{Compassion/empathy} (S~1) is the easiest dimension overall (AUROC 0.87--0.96; EER down to 0.06), likely reflecting that supportive intent is frequently marked by overt acoustic and lexical markers such as prosody and explicit validation, producing clean pole separation in Figure~\ref{fig:violin_plot}. Human correlation is high (0.56--0.97), indicating most judges align with human consensus on empathy strength, with remaining spread plausibly reflecting different weighting of prosody versus wording.

\textbf{Politeness/respect} (S~2) is also strongly separable (AUROC 0.85--0.94). Politeness often appears via explicit courtesies, hedging, and reduced intensity. However, politeness can be strategic (the deliberate use of polite language for an impolite purpose) and markers can appear in both poles, leaving nontrivial overlap even under the optimal decision rule. Figure~\ref{fig:violin_plot} shows higher mid-range density compared to S~1 in both poles. Human correlation is generally high (0.52--0.90), suggesting judges broadly match human rankings but lower values may reflect missed strategic rudeness.

\textbf{Assertiveness} (S~3) remains separable (AUROC 0.73--0.88) but less uniform as prominence and control can reflect either confident guidance or aggression. In Figure~\ref{fig:violin_plot}, negative-pole instances cluster more tightly (consistently subdued delivery), whereas positive-pole scores spread across $[-1,1]$ because ``assertive'' roles include both firm and softly decisive styles. Human correlation is moderate (0.47--0.69), suggesting that some judges may map loudness, pitch range, or sharp timing to “more assertive,” whereas humans may penalize these cues as aggression or impatience.

\textbf{Honesty} (S~4) is the most challenging of S~0--5 (AUROC 0.50--0.72). Honesty may not be a local acoustic stance but a claim about veracity and cross-turn consistency, so short segments provide weak evidence and may encourage default-to-sincere scoring; correspondingly, the negative-pole violins show upward shift and overlap, and pole-order sensitivity reaches 0.29. Human correlation is highly variable (0.08--0.88), suggesting some judges may struggle to match human honesty ratings when the evidence is indirect or depends on cross-turn discourse cues rather than local wording or prosody.

\textbf{Attentiveness} (S~5) is also moderately separable (AUROC 0.76--0.92). Attentiveness can be detected via timing, acknowledgments, and contingent replies, but ``inattentive'' can resemble quiet active listening, yielding an upward-shifted \texttt{N} pole in Figure~\ref{fig:violin_plot} and gaps between pole-consistency (0.76--0.88) and oracle F1 (0.82--0.89). Human correlation is low ($-0.003$--0.40), suggesting judges and humans may rely on different proxies for attention and that models often conflate quiet listening, hesitation, and distraction. This low correlation may also reflect limited human coverage and higher annotator noise for S~5.

Interaction-based stance depends on interpreting the TARGET relative to partner context, increasing ambiguity and framing sensitivity in judge-based evaluation \cite{li2025generation, wang2024large}.

\textbf{Social engagement} (S~6) is moderately separable (AUROC 0.67--0.73) yet poorly thresholded (pole-consistency 0.37--0.64). Figure~\ref{fig:violin_plot} shows substantial overlap and upward shift in the \texttt{N} pole as low engagement can look like brief turns rather than explicit withdrawal. Human correlation is moderate-high (0.40--0.85), implying humans may reward reciprocity and expressiveness more consistently than some judges that underweight interactional context and treat sparse talk as neutral.

\textbf{Power orientation} (S~7) shows the widest dispersion (AUROC 0.56--0.86). Dominance cues (interruptive timing, directives) are salient when present, but accommodation can be realized as ordinary politeness, yielding asymmetric, heavy-tailed distributions in Figure~\ref{fig:violin_plot}. Human correlation varies (0.29--0.82), suggesting that model-human alignment is unstable for this dimension and may depend on how turn-taking, intensity, and lexical directness are weighted.

\textbf{Conflict regulation} (S~8) is moderately separable (AUROC 0.62--0.84) but highly contextual: explicit aggression yields strong negative scores, while many ``conflict'' segments remain near-neutral and overlap with calm regulation, contributing to high disagreement (up to 0.22) and gaps between categorical consistency (0.43--0.70) and oracle F1 (0.58--0.69). Human correlation is high (0.56--0.97), consistent with many judges tracking human rankings of hostility, with remaining spread reflecting difficulty distinguishing calm regulation from low-evidence, near-neutral segments.

\subsection{Model-level comparison, open vs closed, and end-to-end vs cascaded}
GPT (commercial, audio-native) shows the lowest failure rates (0.00--0.005), achieves the best AUROC on 5/9 dimensions (S~0,2,3,4,5), and remains competitive on power orientation (S~7: 0.79), but it has higher pole-order sensitivity on S~7 (0.18). Gemini (commercial, multimodal) is best on compassion and power orientation (S~1: AUROC 0.96, EER 0.06; S~7: AUROC 0.86, EER 0.23) with near-zero failures, but it is weaker on focus/attentiveness (S~5: 0.76) and less separable on warmth and honesty than GPT (Figure~\ref{fig:violin_plot}).

Among open-weight judges, Qwen (end-to-end omni) is the most pole-order stable (S~3/5 disagreement 0.01) yet has higher failure rates (0.03--0.30), upward-shifted \texttt{N} distributions on socially desirable traits (S~0: 0.66; S~4: 0.64), weaker interaction-based separability (S~7: 0.56; S~8: 0.62), and high conflict-regulation sensitivity (S~8 disagreement 0.22). Together, these issues suggest a need for retries, schema enforcement, and bias monitoring. Kimi (end-to-end audio-specialized) is strong on S~0 (pole-consistency 0.77, EER 0.23, AUROC 0.79), S~5 (AUROC 0.87), and the best-tied AUROC on S~6 (0.73), but it is least robust operationally (failures 0.01--0.44) and more order-sensitive (S~0: 0.19; S~4: 0.17). Granite (cascaded, transcript-centric) has near-zero failures and is competitive on S~1 (AUROC 0.87), S~2 (0.85), and S~6 (pole-consistency 0.64; AUROC 0.73), but honesty collapses (S~4 AUROC 0.50) with high pole-order sensitivity (0.29), highlighting the limits of transcript-only judging for pragmatic, delivery-dependent traits.

\subsection{Ablation results}
Ablation results are computed on the subset where both audio- and transcript-based Qwen produced valid structured outputs, ensuring matched evaluated instances.

\textbf{Category 1:} Removing audio cues and restricting the judge to ASR transcripts reduces Qwen’s separability on most single-speaker traits (S~0--S~5), with the largest degradation on sincerity/honesty (S~4 AUROC 0.71 $\rightarrow$ 0.63). Transcript-only decreases AUROC across S~0--S~5 (e.g., S~0 0.74$\rightarrow$0.69; S~5 0.73$\rightarrow$0.70), with smaller but consistent drops on empathy and politeness (S~1 0.85$\rightarrow$0.82; S~2 0.91$\rightarrow$0.87). Human alignment generally benefits from audio as correlation drops under transcript-only on most dimensions (e.g., S~0 0.81$\rightarrow$0.24). This suggests that Qwen’s audio channel provides useful paralinguistic signal in terms of stance reference.

\textbf{Category 2:} For S~6 and S~7, acoustics generally help: removing audio reduces AUROC (S~6 0.68$\rightarrow$0.65; S~7 0.56$\rightarrow$0.54), aligning with the intuition that interactional stance benefits from paralinguistic evidence such as engagement and prosodic emphasis. In contrast, S~8 behaves differently: transcript-only improves AUROC (0.62$\rightarrow$0.67) and reduces EER (0.46$\rightarrow$0.38), yet separability remains weaker than most single-speaker evaluations. This pattern complements the main experiments where S~8 is Qwen’s most pole-order sensitive dimension (disagreement 0.22), suggesting that the remaining S~8 errors are not primarily due to missing acoustic cues, but are more plausibly driven by interaction ambiguity, weak-label mismatch, or insufficient contextual evidence for the conflict-regulation construct in this dataset.

\subsection{Limitations}

Our study has some limitations. This includes weak supervision from role prompts, evaluation on a sampled subset, short segments that can underspecify long-range interactional phenomena, and limited human evaluation coverage. We also note misuse risk: this framework could be misapplied to infer personality traits from voice without consent; our results do not validate stable individual-level trait inference. Despite these constraints, StanceBench offers a concrete, reproducible starting point for stance-centric evaluation that moves ``beyond words'' in spoken interaction.

% \subsection{Practical guidance for judge selection}
% Because StanceBench measures robustness, bias sensitivity, and separability, judge choice can be aligned with user priorities rather than treated as one-size-fits-all \cite{bavaresco2025llms,wang2024fair}.

% For maximum interpersonal stance awareness with minimal engineering overhead, use \textbf{GPT} by default: it has the strongest overall AUROC profile with near-zero failures. For evaluations centered on compassion/empathy (S~1) or power orientation (S~8), \textbf{Gemini} is a competitive alternative and the top performer on those dimensions.

% For open weights and local, reproducible deployment, start with \textbf{Granite} for reliable structured outputs, accepting moderate separability on some traits, and loss of acoustic information. \textbf{Qwen} is most stable, but requires mitigation for high structured-output failure rates and monitoring for positivity skew on warmth and honesty. \textbf{Kimi} is effective on warmth and focus but too failure-prone. Guardrails are required to increase stability (retries, schema enforcement, post-hoc consistency filtering).

\definecolor{green2}{RGB}{147, 215, 168}
\definecolor{blue2}{RGB}{185, 225, 255}

\begin{table}[!t]
\centering
\caption{\small
% Evaluated instances per dimension: S~0=134, S~1=127, S~2=132, S~3=91, S~4=135, S~5=69, S~6=209, S~7=233, S~8=231.
$n$ in each stance label indicates the number of evaluated instances.
Feature names: Dis. rate=mean segment disagreement rate; Pole const.=categorical pole-consistency; Oracle F1=oracle upper-bound F1.
\colorbox{blue2}{Blue} and \colorbox{green2}{Green} highlights the best values for \colorbox{blue2}{robustness} and \colorbox{green2}{separability} per stance.
}
\vspace{-3mm}
\scriptsize
\renewcommand{\arraystretch}{1}
\resizebox{1.01\linewidth}{!}{
\begin{tabular}{p{.2mm} l c c c c c c c}
\toprule
& 
\multirow{3}{*}{\textbf{Model}} &
\textbf{Fail.} &
\textbf{Dis.} &
\textbf{Pole} &
\textbf{Oracle} &
\multirow{2}{*}{\textbf{EER}} &
\multirow{2}{*}{\textbf{AUROC}} &
\textbf{Human} \\
& &
\textbf{rate} &
\textbf{rate} &
\textbf{const.} &
\textbf{F1} &
&
&
\textbf{corr.} \\
& & $\downarrow$ & $\downarrow$ & 
$\uparrow$ & $\uparrow$ & $\downarrow$ & $\uparrow$ & $\uparrow$ \\
\midrule

\multirow{5}{*}{\rotatebox[origin=m]{-90}{\textbf{S~0} (n=134)}} &
Qwen    & 0.15 & {\cellcolor{blue2}0.03} & 0.67 & 0.77 & 0.41 & 0.66 & 0.53 \\
                  & Kimi     & 0.30 & 0.19 & {\cellcolor{green2}0.77} & 0.81 & {\cellcolor{green2}0.23} & 0.79 & 0.70 \\
                  & Granite  & 0.03 & 0.05 & 0.69 & 0.76 & 0.30 & 0.73 & 0.53 \\
                  & GPT      & {\cellcolor{blue2}0.00} & 0.04 & 0.75 & {\cellcolor{green2}0.82} & 0.27 & {\cellcolor{green2}0.81} & {\cellcolor{green2}0.80} \\
                  & Gemini   & 0.01 & 0.05 & 0.70 & 0.80 & 0.27 & 0.75 & 0.64 \\
\midrule

\multirow{5}{*}{\rotatebox[origin=m]{-90}{\textbf{S~1} (n=127)}} & Qwen    & 0.27 & 0.06 & 0.87 & 0.85 & 0.13 & 0.92 & 0.73 \\
                  & Kimi     & 0.32 & 0.07 & 0.89 & 0.87 & 0.13 & 0.93 & {\cellcolor{green2}0.97} \\
                  & Granite  & 0.01 & 0.05 & 0.84 & 0.79 & 0.17 & 0.87 & 0.81 \\
                  & GPT      & {\cellcolor{blue2}0.00} & {\cellcolor{blue2}0.02} & 0.88 & 0.89 & 0.09 & 0.94 & 0.56 \\
                  & Gemini   & 0.02 & 0.04 & {\cellcolor{green2}0.93} & {\cellcolor{green2}0.93} & {\cellcolor{green2}0.06} & {\cellcolor{green2}0.96} & 0.86 \\
\midrule

\multirow{5}{*}{\rotatebox[origin=m]{-90}{\textbf{S~2} (n=132)}} & Qwen    & 0.18 & {\cellcolor{blue2}0.03} & 0.86 & 0.89 & 0.22 & 0.87 & 0.87 \\
                  & Kimi     & 0.21 & 0.04 & 0.87 & 0.90 & 0.15 & 0.87 & 0.52 \\
                  & Granite  & 0.02 & 0.08 & 0.81 & 0.88 & 0.26 & 0.85 & 0.85 \\
                  & GPT      & {\cellcolor{blue2}0.00} & {\cellcolor{blue2}0.03} & {\cellcolor{green2}0.88} & {\cellcolor{green2}0.91} & {\cellcolor{green2}0.12} & {\cellcolor{green2}0.94} & 0.81 \\
                  & Gemini   & 0.010 & {\cellcolor{blue2}0.03} & {\cellcolor{green2}0.88} & {\cellcolor{green2}0.91} & 0.15 & 0.91 & {\cellcolor{green2}0.90} \\
\midrule

\multirow{5}{*}{\rotatebox[origin=m]{-90}{\textbf{S~3} (n=91)}} & Qwen    & 0.15 & {\cellcolor{blue2}0.01} & 0.76 & 0.78 & 0.22 & 0.80 & 0.54 \\
                  & Kimi     & 0.44 & 0.03 & 0.72 & 0.73 & 0.37 & 0.73 & 0.47 \\
                  & Granite  & 0.05 & 0.04 & 0.68 & 0.76 & 0.27 & 0.75 & 0.65 \\
                  & GPT      & {\cellcolor{blue2}0.00} & 0.03 & 0.75 & 0.85 & 0.18 & {\cellcolor{green2}0.88} & {\cellcolor{green2}0.69} \\
                  & Gemini   & 0.005 & 0.03 & {\cellcolor{green2}0.82} & {\cellcolor{green2}0.86} & {\cellcolor{green2}0.17} & 0.87 & 0.62 \\
\midrule

\multirow{5}{*}{\rotatebox[origin=m]{-90}{\textbf{S~4} (n=135)}} & Qwen    & 0.28 & 0.06 & 0.57 & 0.68 & 0.41 & 0.64 & 0.73 \\
                  & Kimi     & 0.30 & 0.17 & 0.62 & 0.67 & 0.37 & 0.67 & 0.71 \\
                  & Granite  & 0.01 & 0.29 & 0.55 & 0.57 & 0.51 & 0.50 & 0.08 \\
                  & GPT      & {\cellcolor{blue2}0.004} & 0.11 & {\cellcolor{green2}0.64} & {\cellcolor{green2}0.72} & {\cellcolor{green2}0.35} & {\cellcolor{green2}0.72} & 0.67 \\
                  & Gemini   & 0.02 & {\cellcolor{blue2}0.05} & 0.59 & 0.70 & 0.37 & 0.66 & {\cellcolor{green2}0.88} \\
\midrule

\multirow{5}{*}{\rotatebox[origin=m]{-90}{\textbf{S~5} (n=69)}} & Qwen    & 0.30 & {\cellcolor{blue2}0.01} & 0.76 & 0.82 & 0.29 & 0.80 & 0.22 \\
                  & Kimi     & 0.38 & 0.06 & 0.83 & 0.85 & 0.17 & 0.87 & {\cellcolor{green2}0.40} \\
                  & Granite  & 0.02 & 0.06 & 0.77 & 0.84 & 0.24 & 0.81 & -0.003 \\
                  & GPT      & {\cellcolor{blue2}0.00} & 0.07 & {\cellcolor{green2}0.88} & {\cellcolor{green2}0.89} & {\cellcolor{green2}0.13} & {\cellcolor{green2}0.92} & 0.36 \\
                  & Gemini   & {\cellcolor{blue2}0.00} & 0.17 & 0.82 & 0.89 & 0.34 & 0.76 & 0.05 \\
\midrule
\midrule

\multirow{5}{*}{\rotatebox[origin=m]{-90}{\textbf{S~6} (n=209)}} & Qwen    & 0.03 & {\cellcolor{blue2}0.05} & 0.37 & 0.56 & 0.37 & 0.69 & 0.40 \\
                  & Kimi     & 0.01 & 0.09 & 0.52 & 0.63 & {\cellcolor{green2}0.31} & {\cellcolor{green2}0.73} & {\cellcolor{green2}0.85} \\
                  & Granite  & {\cellcolor{blue2}0.00} & 0.07 & {\cellcolor{green2}0.64} & 0.59 & 0.36 & 0.73 & 0.66 \\
                  & GPT      & 0.005 & 0.11 & 0.46 & 0.60 & 0.37 & 0.71 & 0.67 \\
                  & Gemini   & {\cellcolor{blue2}0.00} & 0.09 & 0.37 & 0.57 & 0.38 & 0.67 & 0.57 \\
\midrule

\multirow{5}{*}{\rotatebox[origin=m]{-90}{\textbf{S~7} (n=233)}} & Qwen    & 0.05 & {\cellcolor{blue2}0.04} & 0.53 & 0.71 & 0.48 & 0.56 & 0.29 \\
                  & Kimi     & 0.02 & 0.08 & 0.58 & 0.64 & 0.39 & 0.64 & {\cellcolor{green2}0.82} \\
                  & Granite  & {\cellcolor{blue2}0.00} & 0.10 & 0.55 & 0.68 & 0.41 & 0.60 & 0.58 \\
                  & GPT      & {\cellcolor{blue2}0.00} & 0.18 & 0.75 & 0.79 & 0.26 & 0.79 & 0.67 \\
                  & Gemini   & {\cellcolor{blue2}0.00} & 0.11 & {\cellcolor{green2}0.80} & {\cellcolor{green2}0.80} & {\cellcolor{green2}0.23} & {\cellcolor{green2}0.86} & 0.74 \\
\midrule

\multirow{5}{*}{\rotatebox[origin=m]{-90}{\textbf{S~8} (n=231)}} & Qwen    & 0.06 & 0.22 & 0.43 & 0.58 & 0.46 & 0.62 & 0.56 \\
                  & Kimi     & 0.02 & {\cellcolor{blue2}0.06} & {\cellcolor{green2}0.70} & 0.58 & 0.30 & 0.75 & 0.77 \\
                  & Granite  & {\cellcolor{blue2}0.00} & 0.08 & 0.66 & 0.61 & 0.29 & 0.78 & {\cellcolor{green2}0.97} \\
                  & GPT      & {\cellcolor{blue2}0.00} & 0.11 & 0.52 & 0.63 & 0.32 & 0.76 & 0.84 \\
                  & Gemini   & {\cellcolor{blue2}0.00} & 0.10 & 0.68 & {\cellcolor{green2}0.69} & {\cellcolor{green2}0.23} & {\cellcolor{green2}0.84} & 0.82 \\
\bottomrule
\end{tabular}}

\label{tab:judge_metrics_by_Q}
\vspace{-2em}
\end{table}

\section{Conclusion and Future work}
This paper introduced StanceBench, a framework for evaluating interpersonal stance in conversational speech, motivated by the mismatch between rapidly improving speech-to-speech dialogue models and evaluation protocols that still prioritize content fidelity over interactional intent. We (i) operationalized stance as 9 role-prompt pole dimensions (warmth, compassion/empathy, politeness/respect, assertiveness, honesty, attentiveness, social engagement, power orientation, conflict regulation) using the Improvised subset of Seamless Interaction dataset, (ii) standardized single-speaker and interaction-based judging settings, and (iii) reported diagnostics for judge reliability and stance separability, including human correlation on a small human-evaluated subset.

Across the evaluated judges, separability shows stable dimension-level patterns. Compassion/empathy and politeness/respect are consistently the easiest dimensions, with generally strong alignment with human ratings. Warmth is only moderately separable, with positivity skew and upward drift of negative-pole scores, consistent with socially normative mild-positive speech. Assertiveness is separable but asymmetric as high-assertiveness can span calm firmness to confrontation. Honesty is the hardest single-speaker dimension, likely because it depends on indirect discourse cues and cross-turn coherence that are weak in short segments, which also increases pole-order sensitivity and produces variable human correlation. Attentiveness is separable by ranking but shows low human correlation, consistent with the ambiguity between quiet active listening and disengagement. For interaction-based stances, social engagement is moderately separable yet poorly thresholded, power orientation is highly variable, and conflict regulation is most context-sensitive with higher disagreement and near-neutral overlap. The audio vs.\ transcript-only ablation suggests acoustic evidence often helps for interactional stances, but the required evidence type is stance-dependent.

Future work will extend the stance inventory with additional dimensions and roles from the Seamless Interaction dataset and scale human evaluation with more annotators, stratified coverage, and explicit inter-annotator agreement. We will also explore richer interaction-based judging using longer conversational context and turn-taking cues. In parallel, we will develop the benchmark infrastructure as a modular, open evaluation framework with standardized interfaces, enabling researchers to readily integrate and evaluate new judge models within the StanceBench pipeline.

\clearpage

% \raggedbottom

\section{Acknowledgment}
This work was funded by Amazon through the JHU+Amazon ai2ai initiative.

\section{Use of Generative AI Disclosure}
Generative AI was used only for language polishing (grammar, clarity, and style). The tool was not used to generate scientific content, experimental results, analyses, conclusions, or any significant part of the manuscript. 

% \balance

\bibliographystyle{IEEEtran}
\bibliography{mybib}

\end{document}